\documentclass[journal]{IEEEtran}
\renewcommand{\baselinestretch}{0.99}

\usepackage{bm}
\usepackage{ragged2e}
\usepackage{multirow}
\usepackage{amssymb}
\usepackage[pdftex]{graphicx}
\usepackage{graphicx}
\usepackage{epstopdf}
\usepackage{epsfig}
\usepackage{algorithmic}
\usepackage[lined,ruled,commentsnumbered]{algorithm2e}

\usepackage{cite}

\usepackage[cmex10]{amsmath}

\usepackage{algorithmic}

\ifCLASSOPTIONcompsoc
\usepackage[caption=false,font=footnotesize,labelfont=sf,textfont=sf]{subfig}
\else
\usepackage[caption=false,font=footnotesize]{subfig}
\fi

\usepackage{fixltx2e}
\usepackage{stfloats}

\ifCLASSOPTIONcaptionsoff
\usepackage[nomarkers]{endfloat}
\let\MYoriglatexcaption\caption
\renewcommand{\caption}[2][\relax]{\MYoriglatexcaption[#2]{#2}}
\fi
\let\MYorigsubfloat\subfloat
\renewcommand{\subfloat}[2][\relax]{\MYorigsubfloat[]{#2}}

\usepackage{float}

\usepackage{tikz,xcolor,hyperref}
\definecolor{lime}{HTML}{A6CE39}
\DeclareRobustCommand{\orcidicon}{%
    \begin{tikzpicture}
    \draw[lime, fill=lime] (0,0) 
    circle [radius=0.16] 
    node[white] {{\fontfamily{qag}\selectfont \tiny ID}};    \draw[white, fill=white] (-0.0625,0.095) 
    circle [radius=0.007];    \end{tikzpicture}
    \hspace{-2mm}}
\foreach \x in {A, ..., Z}{%
    \expandafter\xdef\csname orcid\x\endcsname{\noexpand\href{https://orcid.org/\csname orcidauthor\x\endcsname}{\noexpand\orcidicon}}
    }

\usepackage{url}
\usepackage{pifont}
\usepackage{enumerate}
\newcommand{\tabincell}[2]{
	\begin{tabular}{@{}#1@{}}#2\end{tabular}
}
\usepackage{color}

\begin{document}

\title{LMVD: A Large-Scale Multimodal Vlog Dataset for Depression Detection in the Wild}

\author{Lang~He \orcidL{} $^{*}$,
Kai Chen,
Junnan Zhao,
Yimeng Wang,
Ercheng Pei,
Haifeng Chen,
Jiewei Jiang,
Shiqing Zhang,
Jie Zhang,
Zhongmin Wang,
Tao He,
Prayag Tiwari

\thanks{This work is supported by National Natural Science Foundation of China (grant 62376215), the Open Fund of National Engineering Laboratory for Big Data System Computing Technology (Grant No. SZU-BDSC-OF2024-16), the Humanities and Social Sciences Program of the Ministry of Education (22YJCZH048), the Key Research and Development Project of Shaanxi Province (2024GX-YBXM-137), and the Xi’an University of Posts and Telecommunications Innovation Fund (CXJJYL2023044).}
\thanks{Lang~He, Kai Chen, Junnan Zhao, Yimeng Wang, Ercheng Pei, Jie Zhang, and Zhongmin Wang are with the School of Computer Science and Technology, Xi'an University of Posts and Telecommunications, Xi'an 710121, Shaanxi, China, with the Shaanxi Key Laboratory of Network Data Analysis and Intelligent Processing, Xi'an 710121, Shaanxi, China, and with the Xi'an Key Laboratory of Big Data and Intelligent Computing, Xi'an 710121, Shaanxi, China. (Corresponding author:
Lang He, Prayag~Tiwari, E-mail:langhe@xupt.edu.cn)}
\thanks{Jiewei Jiang is with the School of Electronic Engineering, Xi'an University of Posts and Telecommunications, Xi’an, 710121, China.}
\thanks{Haifeng Chen is with the College of Electronic Information and Artificial Intelligence, Shaanxi University of Science and Technology, Xi’an, 710016, China. }
\thanks{Shiqing Zhang is with the Institute of Intelligent Information Processing, Taizhou University, Taizhou, 318000, Zhejiang, China }
\thanks{Tao He is with the School of Education, Shenzhen University, Shenzhen, 518060, China.}
\thanks{Prayag~Tiwari is with the School of Information Technology, Halmstad University, Halmstad) (E-mail:prayag.tiwari@hh.se)}

\thanks{Manuscript received XXXXXX; revised XXXXXX.}}

\markboth{IEEE Transactions on Affective Computing}%
{Shell \MakeLowercase{\textit{et al.}}: Bare Demo of IEEEtran.cls for IEEE Journals}

\maketitle

\begin{abstract}
Depression can significantly impact many aspects of an individual's life, including their personal and social functioning, academic and work performance, and overall quality of life. Many researchers within the field of affective computing are adopting deep learning technology to explore potential patterns related to the detection of depression. However, because of subjects' privacy protection concerns, that data in this area is still scarce, presenting a challenge for the deep discriminative models used in detecting depression. To navigate these obstacles, a \textbf{\underline{l}}arge-scale \textbf{\underline{m}}ultimodal \textbf{\underline{v}}log \textbf{\underline{d}}ataset (LMVD), for depression recognition in the wild is built. In LMVD, which has 1823 samples with 214 hours of the 1475 participants captured from four multimedia platforms (Sina Weibo, Bilibili, Tiktok, and YouTube). A novel architecture termed MDDformer to learn the non-verbal behaviors of individuals is proposed. Extensive validations are performed on the LMVD dataset, demonstrating superior performance for depression detection. We anticipate that the LMVD will contribute a valuable function to the depression detection community.  
The data and code will released at the link: https://github.com/helang818/LMVD/.

\end{abstract}


%
\IEEEpeerreviewmaketitle

\section{Introduction}
%
%
%
%
\IEEEPARstart{M}{ajor} 
 depression disorder (MDD) has been projected to a primary mental illness by 2030. 
A comprehensive survey and meta-analysis unveiled that from a study of 41,531 individuals, 33.7\% of them experienced depression during the COVID-19 pandemic \cite{salari2020prevalence}. Normally, depression can affect various aspects of daily activities, including the progress of one's careers, studies, and families, etc \cite{he2022deep}. In some severe cases, individuals suffering from depression may contemplate or attempt suicide \cite{hawton2013risk}.  In spite of the many attempts to identify depression, recent findings suggest that the prevalence of depression may be increasing in the  younger individuals group.

Depressed subjects often express several different non-verbal behaviors, e.g., facial expressions, body gestures, and smile. Diagnostic and Statistical Manual of Mental Disorders (DSM-V) outlines symptoms such as agitation (e.g., the inability to sit still, pacing, and hand-wringing) or retardation (e.g., slowed speech and body movements and increased pauses before answering) that may be displayed by depressed subjects. Current approaches of treating depression are mainly based on assessment by clinicians or the description of depressed subjects, both of which may be subjective. With the fast advancement of  computer vision methods, a list of methods are explored to study the depression severity. Present methods of recognising depression can be considered as hand-crafted \cite{valstar2013avec,valstar2014avec,valstar2016avec,he2018automatic} and deep learning-based \cite{zhu2017automated,zhou2018visually,he2021automatic,he2022reducing,zhang2023improved} approaches. 
From the modality perspective, methods for depression recognition can be classified into audiovisual cues, social media data (Weibo, Twitter), and physiological and non-physiological signals \cite{bhadra2022insight} such as skin conductance, electroencephalography (EEG), magnetoencephalography (MEG), electrocardiography (ECG), heart sounds, respiration, and pulse signals. Despite promising performances achieved by current depression recognition methods, several challenges remain in effectively recognizing depression via multimodal signals. 
This is especially prevalent in the field of deep learning community, where a substantial quantity of data samples is necessary to train the available models. By reviewing the depression databases from recent decades \cite{he2022deep}, one observes small samples of publicly accessible data because of privacy protection. Furthermore, most of the available databases are collected in controlled laboratory environments via doctor-patient interactions; therefore behavioral patterns of the depressed subjects outside of the laboratory environment are missed \cite{huang2020domain}. Although, Yoon et al. \cite{yoon2022d} introduced a D-vlog depression dataset containing 961 samples and providing facial landmarks and low level descriptors (LLDs). 

Hence, to mitigate the above-mentioned major issues, a large-scale multimodal vlog database (LMVD) is built. In general, these vlogs are recorded and uploaded spontaneously by users, which motivate us to release this novel dataset to capture the potential patterns embedded in the vlogs of individuals navigating their daily lives. First, we obtain the vlogs from three Chinese multimedia platforms (Sina Weibo, Bilibili, and Tiktok) based on the following keywords related to depression (``depressed category": depression, my depressed life, and depressed vlog) in the depressed category and; "health category:" daily life, daily vlog, my vlog in the Non-depressed category. In addition, for each word, the following labels are also considered such as lower mood, loss of interest in many things, insomnia, and persistent thoughts of death or suicide. Secondly, we ask the volunteers to clean and count the unusable video clips, e.g., gender, duration time, etc. Thirdly, we ask the volunteers and clinicians to annotate the videos. 
In this stage, the volunteers first checked and annotated the quality of video vlogs, while clinicians verified the labels (right or not) and made the final decisions. After this, we decode the audio and Chinese text from the vlogs, and extract the audio features by using the pre-trained VGGish model \cite{abu2016youtube}, and extract the visual features, i.e., facial action units (FAUs), landmarks, eye gaze, and head pose. Finally, we introduced a multimodal depression recognition architecture to fuse the non-verbal behavioral features of the audio and video. In this architecture, a Transformer module to learn the potential characteristics, and an attentional multimodal feature fusion mechanism is proposed to capture the non-verbal patterns from the audiovisual features. To establish a baseline and open the dataset to researchers in the community of affective computing, we adopt both machine learning and deep learning methods; a great number of validations were performed on the collected dataset, obtaining excellent performances. 



The novelties of this study are highlighted as follows:

\begin{enumerate}
    \item To encourage collaboration and to assist with future studies, the LMVD is publicly available\footnote{https://github.com/helang818/LMVD/}. Due to privacy protection, the raw vlogs and detailed information are not accessible. However, to assist the researchers, we provide the following data and features: raw audio signals, the learned features of VGGish, FAUs, landmarks, head pose, and eye gaze features. The proposed dataset has 1823 samples (214 hours) from 1475 participants. After reviewing the current studies on depression detection, we can confirm that our collected dataset is the first study in the field that can be used for depression detection.
    \item To provide a benchmark dataset and generate a baseline in the affective computing community, machine learning and deep learning methods are adopted to perform the experimental validations. In addition, we also leverage the transformer and cross attention mechanism to learn the complementary non-verbal behaviors from the audiovisual features. Using LMVD, 
    we obtain the values of 76.85\%, 76.88\%, 77.02\%, 76.88\% for the F1-score, accuracy, precision, recall, respectively. Moreover, the performances are illustrated to further showcase the effectiveness of MDDformer.    
\end{enumerate}
The remained of the present paper is structured as follows. Section \ref{sec:Related} details previous study on COVID-19 detection. Our method is introduced in Section \ref{sec:Method}. Section \ref{sec:experiments} discusses the experimental performance. Conclusions and future studies are planned in Section \ref{sec:conclusion}.

\section{Related Works}\label{sec:Related}
In this section, we offer a detailed explanation of the collected dataset and its relevance to multimodal depression detection by reviewing related works. 

\subsection{The Depression Dataset}

\begin{table*}[h]
	\centering
	\caption{A review of the depression datasets in the the past thirty years. ``A" denotes audio, ``V" denotes video, and ``T" denotes text. ``Pu" denotes public, ``Pr" represents private.  }
	\begin{tabular}{ccccc}	
		\hline
		\multicolumn{1}{c}{\bf Database} & \multicolumn{1}{c}{\bf Modality} &
		\multicolumn{1}{c}{\bf \tabincell{c}{Subjects}} &		
		\multicolumn{1}{c}{\bf Samples}   &\multicolumn{1}{c}{\bf Pu /Pr}\\
		\hline
1: DementiaBank \cite{becker1994natural}(1994) & A+V+T  &226&    &Pu\\	
2: -- \cite{stassen1998speech}(1998)&A&43& &Pr\\
3: -- \cite{france2000acoustical}(2000)&A&115&&Pr\\
4: -- \cite{alpert2001reflections}(2001)&A&41&&Pr\\
5: -- \cite{moore2004comparing}(2004)&A&33&&Pr\\
6: -- \cite{yingthawornsuk2006objective}(2006)&A&32&&Pr\\
7: -- \cite{cohn2009detecting}(2009)&A&57&&Pr\\
8: ORI \cite{maddage2009video}(2009)& V &139&&Pr   \\
9: BlackDog \cite{Alghowinem2012From}(2009)& A+V &80&&Pr \\
10: ORYGEN \cite{ooi2011prediction}(2011)&V&191& &Pr  \\
11: -- \cite{mundt2012vocal} (2012)&A&165&&Pr\\	
12: AVEC2013 \cite{valstar2013avec} (2013)& A+V&292&150 &Pu (raw data) \\
13: AVEC2014 \cite{valstar2014avec} (2014) & A+V  &292&300  &Pu (raw data) \\
14: Crisis Text Line \cite{chen2014visualizations} (2014) &  T &--&   &Pu (--) \\
15: DAIC-WoZ \cite{gratch2014distress} (2014)&  A+V+T&110&189 &Pu (raw audio and audiovisual features) \\
16: Rochester \cite{zhou2015tackling} (2015)& V&27&  &Pr \\
17: CHI-MEI \cite{huang2016unipolar}(2016)	&V &53&&Pr\\		
18: Pittsburgh \cite{dibekliouglu2018dynamic} (2018)& A+V &49&130  &Pu (audiovisual features) \\
19: MODMA \cite{cai2020modma} (2020) &A+EEG &55& &Pu (raw audio and EEG) \\		
20: CMDC \cite{9793717} (2022) &A+V+T&78&  &Pu (audiovisual features) \\
21: D-vlog \cite{yoon2022d} (2022) & A+V&816&961  &Pu (audiovisual features (LLD, facial landmarks)) \\		
22: Ours (LMVD) (2024) & A+V &1475 & 1823 &Pu (audio, various audiovisual features) \\		
		\hline               
	\end{tabular}
	
	\label{table:database_depression}
\end{table*}

Because of the privacy protection related to studies on depression, data collection is very complicated. Consequently, various research teams have endeavored to record their own databases for depression estimation. This section examines a total of 21 databases, with only nine being accessible to the public. Table \ref{table:database_depression} shows the available audiovisual depression databases over the past 30 years. Since 1994, depression recognition gained attention from researchers, resulting in the release of a dataset by Becker et al \cite{becker1994natural}. 

Based on Table \ref{table:database_depression}, the following observations are considered:

\begin{enumerate}

    \item \textbf{Data Accessibility:} Eight databases on depression detection are publicly available. However, most of the databases provide the extracted hand-crafted and deep-learning features, without providing the raw audio and video signals. Most importantly, our collected dataset has more samples for multimodal depression detection among individuals navigating their daily lives. 
    
    \item \textbf{Sample Size:} In terms of the number of data samples, one can see that only the D-vlog \cite{yoon2022d} holds the previous record with 961 samples. LMVD boasts a significantly larger collection of 1,823 samples (214 hours) from 1,475 participants. 
    However, the databases AVEC2013 and AVEC2014 are the most frequently used by researchers, although AVEC2014 only contains 292 subjects with 300 samples.

    \item \textbf{Geographic Diversity:} Most of the databases are located in the EU, with only two collected in China. Based on the concepts of depression, the depressed subjects represent different behaviors in different countries.
            
    \item \textbf{Modality Coverage:} Most of the datasets contain audio and video modalities, while only a few contain the text modality for depression detection. Similar to existing datasets, LMVD primarily focuses on audio and video modalities. However, it offers a wider range of visual features compared to some existing datasets, including FAUs, landmarks, eye gaze, and head pose features, which can provide richer insights. As indicated in Table \ref{table:database_depression}, 12 databases only have unimodal, and 50\% of the them have the audio modality, because this metric is easily recorded. There are only eight (45\%) multimodal databases available.
	
\end{enumerate}

Overall, LMVD offers several advantages for researchers in the field of multimodal depression detection. Its larger sample size, diverse participant pool, raw data access, and broader range of features make it a valuable resource for advancing research efforts.

\subsection{Multimodal Depression Detection}

A series of studies, which utilize cues from audio, video, and text modalities, have been suggested to accurately evaluate the status of depression.  Lam et al. \cite{lam2019context} leverage topic modelling strategy to augment the size of the data and combine the power of a transformer mechanism with a 1D-CNN (1 dimension convolutional neural networks) to capture the patterns from acoustic features. 
Niu et al. \cite{niu2020multimodal} adopt combined spatio-temporal attention (STA) and 
multimodal attention feature fusion (MAFF) network for modeling the multimodal features. 
In a later study \cite{niu2021hcag}, Ni et al. designed a hierarchical context-aware graph (HCAG) attention model that reflects layered information for the assessment of depression and employed a graph attention network (GAT) to discern contextual connections within the text/audio modalities. 
In the work of \cite{chen2023semi}, a graph neural
network-based semi-supervised domain adaptation (GNN-SDA) technique is presented to address the challenges associated
with limited sample sizes and isolated data clusters. Pan et al. \cite{pan2023integrating} present an audiovisual attention architecture named AVA-DepressNet, which focuses on privacy protection concern and an embedded attention-driven module for identifying depression. Moreover, an adversarial multistage (AMS) approach
is formulated for refining the encoder-decoder framework, integrating knowledge of facial structures. In 2024, a
transformer-based structure was introduced from the video, audio, and remote photoplethysmographic (rPPG) cues for multimodal prediction of depression \cite{fan2024transformer}.


\section{LMVD Depression Dataset}\label{sec:Method}

This section introduces the LMVD dataset, a large-scale multimodal dataset built ``in the wild" for depression detection. We elaborate on the following aspects: (1) the motivation of this study, (2) the procedure of collecting the Chinese vlogs from the four media platforms, (3) the step for annotating the vlogs, (4) the step of prepossessing the details, and (5) the extraction of audio and visual features. 

\subsection{Motivation}

As reported in Table \ref{table:database_depression}, the limitations of the available datasets motivated us to build a large-scale dataset for depression detection in individuals navigating their daily lives. This large-scale dataset offers several advantages:
\begin{enumerate}
    \item \textbf{Promotes Research and Applications:} the collection of LMVD will boost both research endeavors and clinical scenarios in the community of automatic depression detection.
    \item \textbf{Benefits Various Stakeholders:} the developed prototype system offers an efficient solution
    applicable to various sectors, including government agencies, hospitals, and universities.
\end{enumerate}

\subsection{Data Collection}

Our goal is to build a large-scale dataset for multimodal depression detection in individuals navigating their daily lives. To achieve this, we aim to collect depression and non-depression vlogs from different platforms with similar content distribution. Therefore, we collect the vlog videos from three Chinese multimedia platforms (Sina Weibo, Bilibili, and Tiktok). Data was collected from 1st Jan 2019 to 30th October 2023 using the following keywords (in Chinese): 
\begin{enumerate}
    \item \textbf{Depressed category:} depression, my depressed life, and depressed vlog. 
    \item \textbf{Non-depressed category:} daily Life, daily vlog, my vlog.	
\end{enumerate}

In addition to the Chinese platforms, we collect vlogs from YouTube using the same keywords in the same period. The collection process followed a similar approach. By collecting data from both Chinese and English platforms, we aimed to enhance the diversity and generalized ability of the LMVD dataset.

\begin{table*}[!t]
\centering
\caption{The statistics of the collected videos from three Chinese multimedia platforms (Sina Weibo, Bilibili, and Tiktok).}
\begin{tabular}{cccc}
\hline
Platform   & Depressed category & Non-depressed category & All     \\
\hline
Bilibili   & 1091      & 1093   & \textbf{2184} \\
Tiktok     & 1147      & 1050   & \textbf{2197} \\
Sina Weibo & 65        & 214    & \textbf{279}  \\
Total           & \textbf{2303}      & \textbf{2357}   & \textbf{4660} \\
\hline
\end{tabular}
\label{tab:sta1}
\end{table*}

As shown in Table \ref{tab:sta1}, the two platforms have balanced samples except for Sina Weibo. This is because many users adopt Sina Weibo to post text messages about their feelings, encompassing their lives, careers, and emotions. This resulted in 65 and 214 samples for the Depressed and Non-depressed. In total, there are 2303 and 2357 samples for the Depressed and Non-depressed categories, respectively. Bilibili, TikTok, and Sina Weibo have 2184, 2197, and 279 samples,
respectively. In total, 4660 vlogs from the three Chinese multimedia platforms are used. In addition, the vlogs from YouTube are of good quality and are used directly in this work.

\subsection{Data Annotation}

\begin{table*}[!t]
\centering
\caption{The statistics of the cleaned vlog videos from the three Chinese (Sina Weibo, Bilibili, Tiktok) multimedia platforms and YouTube.}
\begin{tabular}{cccc}
\hline
Platform   & Depressed category & Non-depressed category &  All    \\ \hline
Bilibili   & 222          & 334           & \textbf{556}  \\
Tiktok     & 314          & 133           & \textbf{447}  \\
Sina Weibo & 65           & 48            & \textbf{113}  \\
YouTube & 307           & 400            & \textbf{707}  \\
 Total          & \textbf{908} & \textbf{915}  & \textbf{1823} \\
\hline
\end{tabular}
\label{tab:sta2}
\end{table*}

  
To perform the annotation for the depression and health vlogs, four master and ten undergraduate students are recruited. First, we ask the ten undergraduate students to check whether the vlogs have faces and if the audio is synthesized. As shown in Table \ref{tab:sta2}, we obtained 1823 data samples for the LMVD. For the Depressed and Non-depressed categories, the number of vlogs is 908 and 915, respectively. Secondly, we ask the four master’s students to recheck the vlogs to ensure the quality is sufficient
for training the deep models for multimodal depression detection. Then we assign the 1823 vlogs to the ten undergraduate students who assigned them to either the Depressed or Non-depressed categories. Finally, the master students check the labels assigned by the undergraduate students.




\begin{table*}[!t]
\centering
\caption{The durations of the cleaned vlog from the three Chinese (Sina Weibo, Bilibili, Tiktok) and the YouTube multimedia platforms.}
\begin{tabular}{cccc}
\hline
Platform   & Depressed category (s) & Non-depressed category (s) &  All(s)    \\ \hline
Bilibili   & 90296.40  & 168395.92 & \textbf{258692.32}  \\
Tiktok     & 35384.80 & 19344.84 & \textbf{54729.64}  \\
Sina Weibo & 13889.23 & 28966.00 & \textbf{42855.23}  \\
YouTube & 197698.23    & 218654.64  & \textbf{416352.87}  \\
 Total          & \textbf{337268.66} & \textbf{435361.40}  & \textbf{772630.06s(214 hours)} \\
\hline
\end{tabular}
\label{tab:sta4}
\end{table*}

Table \ref{tab:sta4} shows the duration of vlogs from different platforms. Bilibili exhibits the most significantly higher time spent between the Depressed and Non-depressed category. As a matter of fact, it shows the highest disparity between the two categories among all listed platforms. TikTok shows less time spent in both categories compared to Bilibili, but still has a considerable amount of time spent by users on the platform. The ratio of time spent between Depressed and Non-depressed is less than Bilibili, indicating a more balanced usage among the two categories. Sina Weibo displays a trend wherein users in the Non-depressed category spend more than twice the amount of time compared to those in the Depressed category. This implies that Sina Weibo is predominantly utilized by users who are not classified as depressed. YouTube has the closest time spent between the two categories, which might suggest a more uniform distribution of usage across users categorized as Depressed and Non-depressed.

\subsection{Multimodal Feature Extraction}

To establish a foundational benchmark for the field of depression recognition, we employ the primary features commonly used.

For audio features, the pre-trained VGGish \cite{hershey2017cnn} model is adopted. This is because the traditional
hand-crafted features have the following limitations: (1) professional knowledge is often needed to design the discriminative
features, and (2) additional valuable patterns may be lost in developing the deep features.

\begin{figure}[!t]
	\centering
	\includegraphics[scale=0.9]{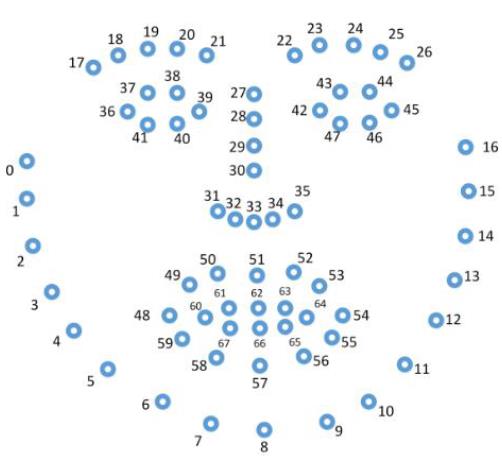}
	\caption{Illustration of the 68 key points for facial landmarks feature. } 
	\label{fig:landmarks}
\end{figure}

For visual features, FAU, facial landmarks, eye gaze, and head pose features are adopted. 

\begin{enumerate}
    \item Facial Action Units (FAUs): We focus on a subset of 17 AUs (AU01, AU02, AU04, ..., AU45) that have been linked to emotional expression. These features represent specific muscle movements in the face that can provide insights into a person's emotional state.
    
    \item Facial Landmarks: We extract facial landmarks (see Fig. \ref{fig:landmarks}) to represent the key points of the participant's facial structure. Facial landmarks are robust for capturing facial muscle movements, making them valuable features for tasks like emotion analysis, depression detection, and facial action unit detection.
    
    \item Eye Gaze: We extract eye gaze features to describe the direction of a participant's gaze. This information is represented by four sets of eye movement feature vectors with 12 dimensions each. The first two sets define the direction of eye movements in the coordinate space, while the latter two sets define the direction based on the head coordinate space. The values ``0" and ``1" represent the left and right eyes, respectively.
    
    \item Head Pose: We extract head pose features to capture the position and rotational direction of the participant's head. These features are represented by a 6-dimensional vector.
\end{enumerate}

\section{Methods}

In this section, we describe the pipeline for multimodal depression detection using the LMVD. We first provide a brief overview of the pipeline, followed by a detailed description of the multimodal depression detection method. 

\subsection{Architecture Overview}

\begin{figure*}[!t]
	\centering
	\includegraphics[scale=0.5]{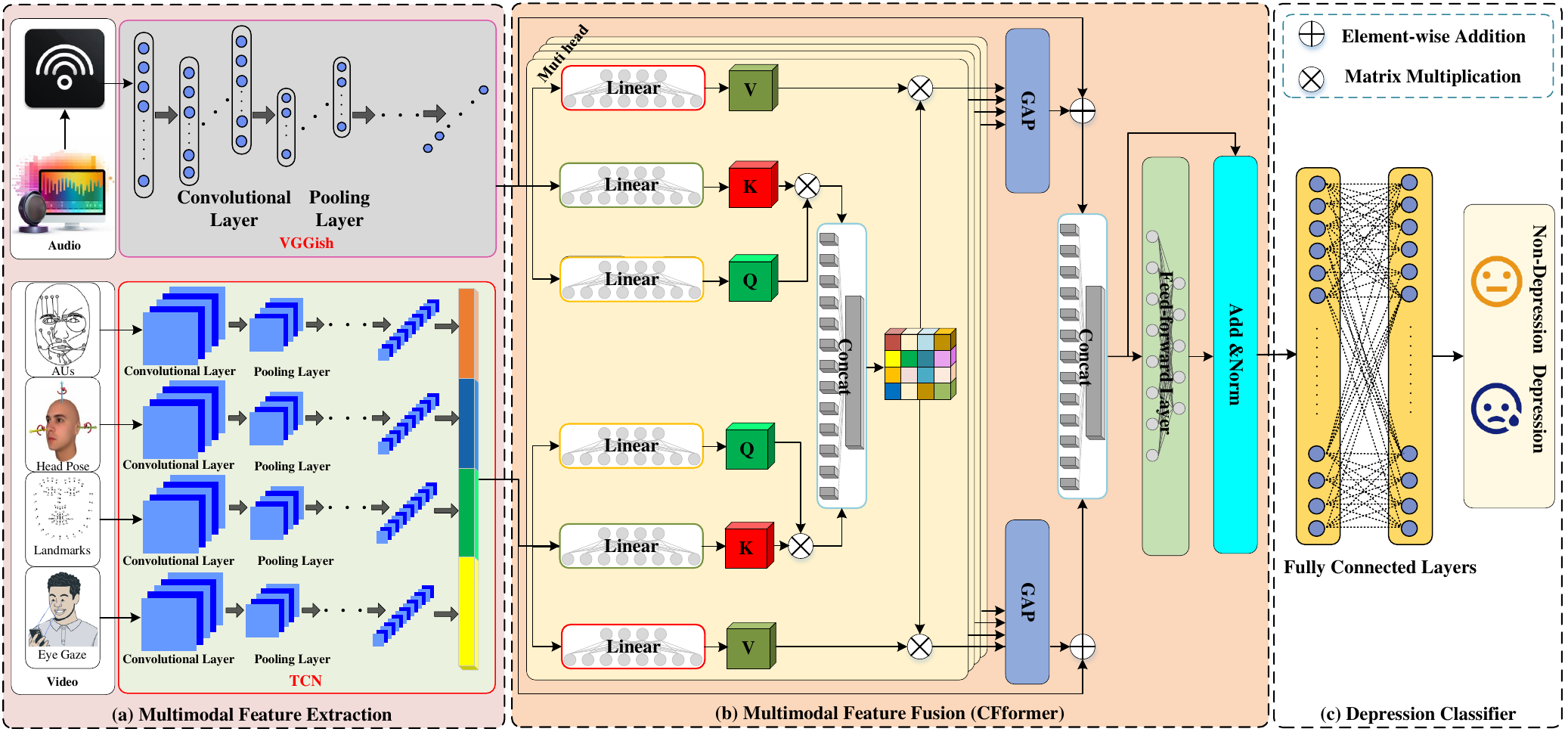}
	\caption{The MDDformer model comprises three main steps: (a) \textbf{Multimodal Feature Extraction:} For the audio cue, the deep features are extracted by VGGish. For the visual cue, AUs, head pose, landmarks, and eye gaze are extract by TCN architecture. (b) \textbf{Multimodal Feature Fusion:} Cross fusion transformer (CFformer) block leverages cross fusion to learn the combined informative behaviors from the audiovisual cues. (c) \textbf{Depression Classifier:} Two fully connected layers and the softmax function are adopted for predicting the depression. } 
	\label{fig:pipeline}
\end{figure*}

Fig. \ref{fig:pipeline} illustrates the proposed MDDformer. First, the audio features are extracted by the VGGish architecture. For the visual cue, AUs, head pose, landmarks, and eye gaze features are extracted by TCN architecture. Then, the CFformer can adopt the advantages of cross fusion for learning behaviors from the audiovisual cues. Finally, two fully connected layers and the softmax function are performed to predicting the depression. 

\subsection{Multimodal Depression Detection}

\subsubsection{Baseline Model}

Over the past decades, deep learning methods have gathered significant attention across various tasks. 
Consequently, to establish
a benchmark in the field of depression detection, we adopt both traditional machine learning methods and several deep
learning techniques. Specifically, K-nearest neighbor (KNN) \cite{peterson2009k} is employed to address two category problems, serving as the
baseline model for multimodal depression detection. More information on KNN can be found in \cite{peterson2009k}.

\subsubsection{Structure of MDDformer}

To effectively detect the discriminative patterns within audiovisual cues, MDDformer is proposed. Let's define the audio feature as $\textbf{X}^a \in \mathbb{R}^{N\times D_a}$, and the video feature as $\textbf{X}^v \in \mathbb{R}^{N\times D_v}$ before inputting into the MDDformer. Here, $D_a$ represents the dimension of audio, $D_v$ represents the dimension of video, and $N$ denotes the length of sequences. 

Initially, the transformer architecture \cite{vaswani2017attention} is proposed to model therelationships in natural language processing (NLP) tasks, which consists of an encoder-decoder structure. Both the encoder and decoder are composed of multiple identical layers, each containing two main sub-modules,i.e., multi-head self-attention and position-wise feed-forward networks. In our task, to fuse the patterns from audio and video branch, an MDDformer is proposed.

The input audio feature $\textbf{X}^a \in \mathbb{R}^{N\times D_a}$ is mapped to three matrices by three linear transformations, i.e., key $K_a$, query $Q_a$, and value $V_a$.
\begin{equation}
Q_a = {\textbf{X}^{\rm{a}}}{W_{Q_a}},K_a = {\textbf{X}^{\rm{a}}}{W_{K_a}},V_a = {\textbf{X}^{\rm{a}}}{W_{V_a}}
\end{equation}
where $W_{Q_a}$, $W_{K_a}$, and $W_{V_a}$ denotes the weights of linear transformation. 
The video feature $\textbf{X}^v \in \mathbb{R}^{N\times D_v}$ can be mapped to three matrices by three linear transformations, i.e., key $K_v$, query $Q_v$, value $V_v$.
\begin{equation}
Q_v = {\textbf{X}^{\rm{v}}}{W_{Q_v}},K_v = {\textbf{X}^{\rm{v}}}{W_{K_v}},V_v = {\textbf{X}^{\rm{v}}}{W_{V_v}}
\end{equation}
where $W_{Q_v}$, $W_{K_v}$, and $W_{V_v}$ denotes the weights of linear transformation.

Next, $Q_a$ is multiplied with $K_a^T$ to generate the feature $F_a = Q_a K_a^T$ and $Q_v$ multiplied with $K_v^T$ to generate the feature $F_v = Q_v K_v^T$. Then we concatenate $F_v$ and $F_a$ to generate the feature map, i.e., $F_{av}$:  
\begin{equation}
F_{av} = concat(F_a,F_v)
\end{equation}

The self attention of the audio branch can be expressed as:
\begin{equation}
\text { Attention\_a }(Q_a, K_a, V_a)=\operatorname{softmax}\left(\frac{F_{av}}{\sqrt{d_k}}\right) V_a
\end{equation}
where $d_k$ is the dimension of the $F_{av}$ matrix. 

The self attention of the visual branch can be expressed as:
\begin{equation}
\text { Attention\_v }(Q_v, K_v, V_v)=\operatorname{softmax}\left(\frac{F_{av}}{\sqrt{d_k}}\right) V_v
\end{equation}
where $d_k$ is the dimension of the $F_{av}$ matrices. 

Then, we concatenate the outputs of each head and reshape them to add with the feature $\textbf{X}^a \in \mathbb{R}^{N\times D_a}$ ,generating the fusion feature.
\begin{equation}
\begin{array}{l}
{F_f}{\rm{ = }}Concat((Concat\left( {} \right.hea{d_{{1_a}}},hea{d_{{2_a}}}, \ldots ,hea{d_{{h_a}}}){W_a}\\
 + {{\bf{X}}^a}){\rm{ + }}(Concat\left( {} \right.hea{d_{{1_v}}},hea{d_{{2_v}}}, \ldots ,hea{d_{{h_v}}}){W_v} + {{\bf{X}}^v}))
\end{array}
\end{equation}
where $h$ is the heads of the multi-head self-attention and $W_a$ and $W_v$ are weight matrices. 

Following the concatenate operation, $F_f$ is then input to the feed-forward network, adopting the add/norm operation to generate the feature $F_{n}$:
\begin{equation}
    F_{n} = Norm(FFN(F_f) + F_f)
\end{equation}
where $Norm$ is the normalization operation.

Following the add and layer normalization, two fully connected layers with ELU activation, and dropout operation are performed on $F_{n}$, represented by $F_{p}$:
\begin{equation}
F_{p}=Softmax(FC(ELU(Dropout)))
\end{equation}
where $Dropout$, $ELU$, $FC$, and $Softmax$ represent dropout, activation, fully connected layer, and softmax function, respectively. 

Finally, cross entropy mechanism is adopted as the loss function:
\begin{equation}
L = {\textstyle{1 \over N}}\sum\nolimits_i {}  - [{y_i} \cdot \log ({p_i}) + (1 - {y_i}) \cdot \log (1 - {p_i})]
\end{equation}
where $N$ denotes the list of samples, $y_i$ is the label (depression and non-depression), and $p_i$ is the predicted value (between 0 and 1). 

\section{Results}\label{sec:experiments}

We elaborate on the details the experimental setup and describe the MDDformer performances in this section. 

\subsection{Experimental Setup}\label{sec:Setup}
We implement and trained the MDDformer model using the PyTorch deep learning toolkit. A 10-fold cross-validation is performed for validating the efficiency of MDDformer. The Adam optimizer is set to $\beta=(0.9,0.999)$ and $\epsilon=1e8$ with the batch size of 4. The initial training learning rate size is 0.00001 and then updated with CosineAnnealingLR decay.
To overcome the overfitting problem, a dropout of 0.2 is adopted in the linear layers and the total epochs is set to 300. Our architecture is evaluated on four NVIDIA Tesla V100-DGX with 32GB.

\subsection{Evaluation Metrics}

In general, the classification performance is mainly evaluated using five metrics for binary classification problems: accuracy, precision, recall (also known as 
sensitivity), specificity, and F1-score. Here, true positive (TP) indicates samples with positive labels that are correctly predicted as positive. Similarly, true negative (TN), false positive (FP), and false negative (FN) respectively represent samples correctly predicted as negative, incorrectly predicted as positive, and incorrectly predicted as negative, respectively. The formula can be expressed as: 

\begin{equation}
Accuracy = \frac{{TP + TN}}{{TP + FP + TN + FN}}
\end{equation}

\begin{equation}
Precision = \frac{{TP}}{{TP + FP}}
\end{equation}

\begin{equation}
Recall = \frac{{TP}}{{TP + FN}}
\end{equation}

\begin{equation}
F1-score = 2 \times \frac{{Precision \times Recall}}{{Precision + Recall}}
\end{equation}

\subsubsection{Performances of Baseline Methods}

\begin{table*}[]
\centering
\caption{Performance of the different baseline methods and the MDDformer. Accuracy, precision, recall, and F1-score are
adopted as the evaluation metrics for depression detection.}
\begin{tabular}{ccccc}
			 			
\hline
Model& Accuracy(\%) & Precision(\%) & Recall(\%) & F1-score(\%) \\
\hline
KNN              & 58.34     & 59.75      & 58.34   & 56.87     \\
SVM              & 64.66     & 65.77      & 64.66   & 64.06     \\
LR               & 64.88     & 65.19      & 64.88   & 64.73     \\
RF               & 69.23     & 69.34      & 69.23  & 69.17     \\
Xception(add)    & 70.99     & 71.92      & 70.99   & 70.67     \\
ViT(add)         & 71.16     & 71.91      & 71.16   & 70.92     \\
Xception(concat) & 71.38     & 71.94      & 71.38   & 71.19     \\
BiLSTM(add)      & 72.31     & 72.69      & 72.31   & 72.20     \\
SEResnet(concat) & 72.54     & 73.13      & 72.54   & 72.36  \\
BiLSTM(concat)   & 72.59     & 73.02      & 72.59   & 72.47     \\
SEResnet(add) & 72.92     & 73.71      & 72.92   & 72.69     \\
ViT(concat)      & 73.03     & 73.52      & 73.03   & 72.90     \\
MDDformer        & \textbf{76.88}     & \textbf{77.02}      & \textbf{76.88}   & \textbf{76.85}  \\
\hline
\end{tabular}
\label{tab:results}
\end{table*}

To further validate the performances of the MDDformer, several machine learning and deep learning architectures are adopted i.e., KNN, SVM, LR, RF, Xception, ViT, BiLSTM, and SEResnet. To make a fair comparison, the weighted accuracy, precision, recall, and F1-score are adopted. As shown in Table \ref{tab:results}, the MDDformer obtains the best performance in term of the evaluation metrics. The terms ``add" and ``concat" represent the addition and concatenate operation, respectively. In our task, we list the models in ascending order according of accuracy. One
can note that the accuracy values of 58.34\%, 64.66\%, 64.88\%, 69.23\%, 70.99\%, 71.16\%, 71.38\%, 72.31\%, 72.54\%, 72.59\%,
72.92\%, 73.03\%, 76.88\% (see Table \ref{tab:results}).

\begin{figure*}[!t]
	\centering
	\includegraphics[scale=0.65]{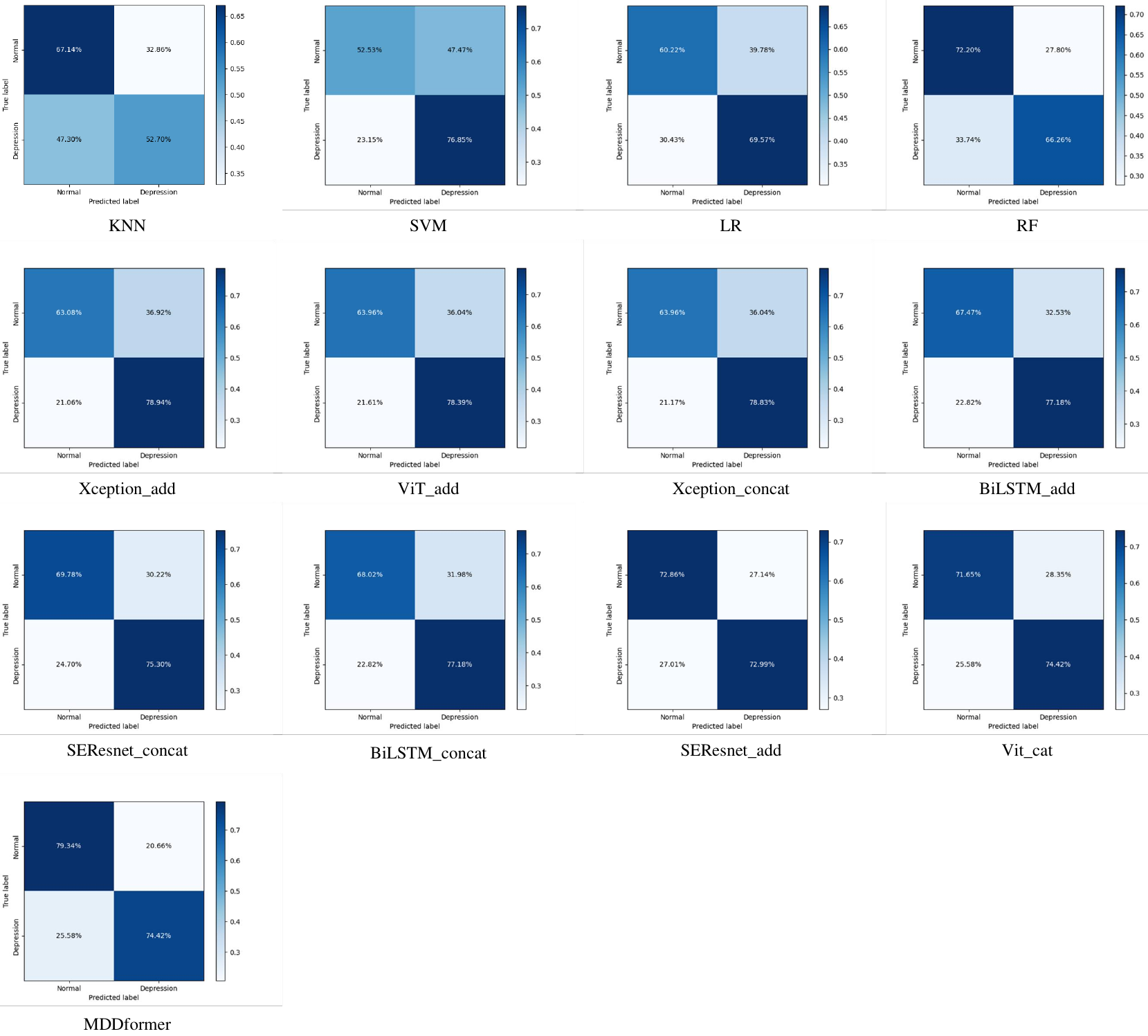}
	\caption{Confusion matrix of the MDDformer and other baseline methods. Each row represents the true labels, and each column represents the predicted values. Element $(m, n)$ indicates the percentage of samples from class $m$ being classified as class $n$.} 
	\label{fig:confusion}
\end{figure*}

Fig. \ref{fig:confusion} provides the confusion matrix for the MDDformer with other baseline architectures. The MDDformer obtains the best performances regarding the classification of depression and non-depression. Note that each row represents the true labels, and each column represents the predicted values. 

\begin{figure*}[!t]
	\centering
	\includegraphics[scale=0.6]{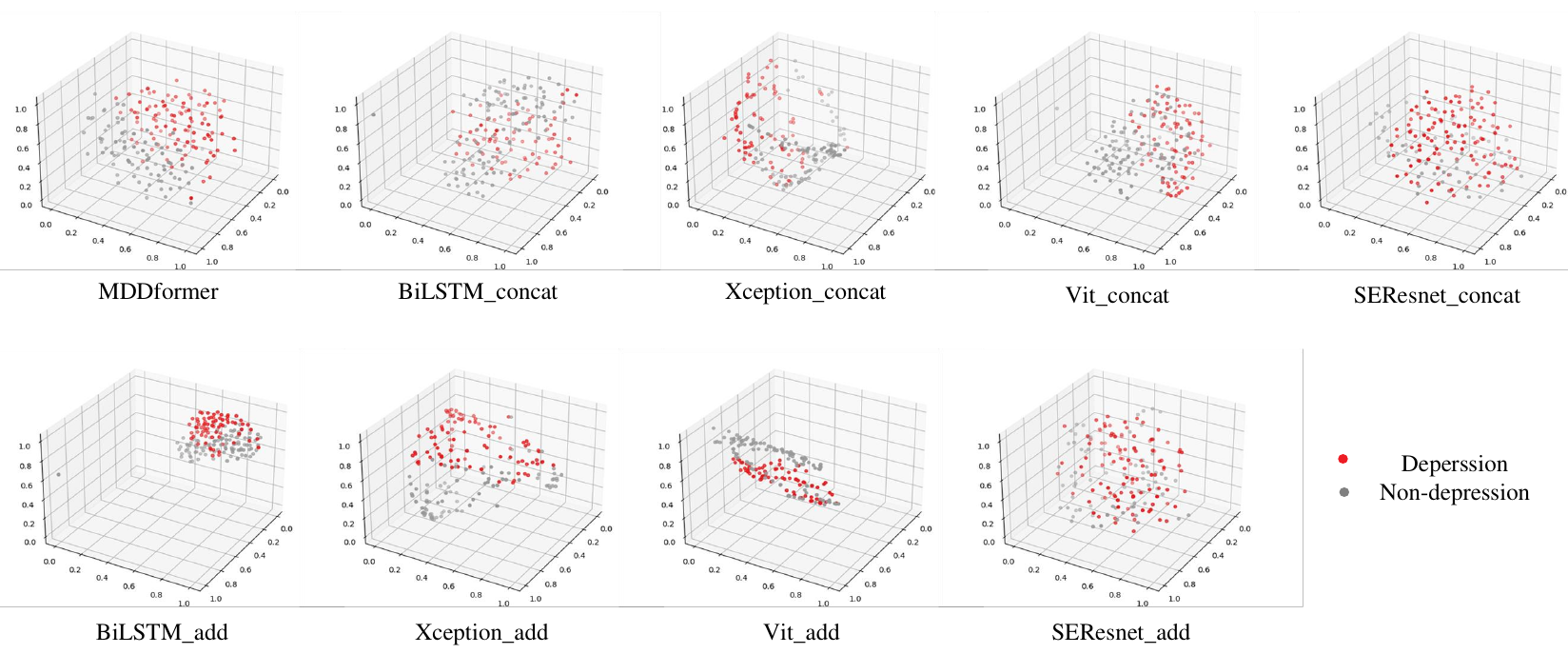}
	\caption{Visualisation of the multimodal features using 3D t-SNE. The red dots represent data from depressed subjects, while the gray dots represent data from healthy control subjects. } 
	\label{fig:tsne}
\end{figure*}

From Fig. \ref{fig:tsne}, it is evident that the classification results of the MDDformer stand out compared to those of other models. The MDDformer yields clearer and more uniform results, indicating its superior performance in classification effectiveness over other methods.

\begin{figure}[!t]
	\centering
	\includegraphics[scale=0.45]{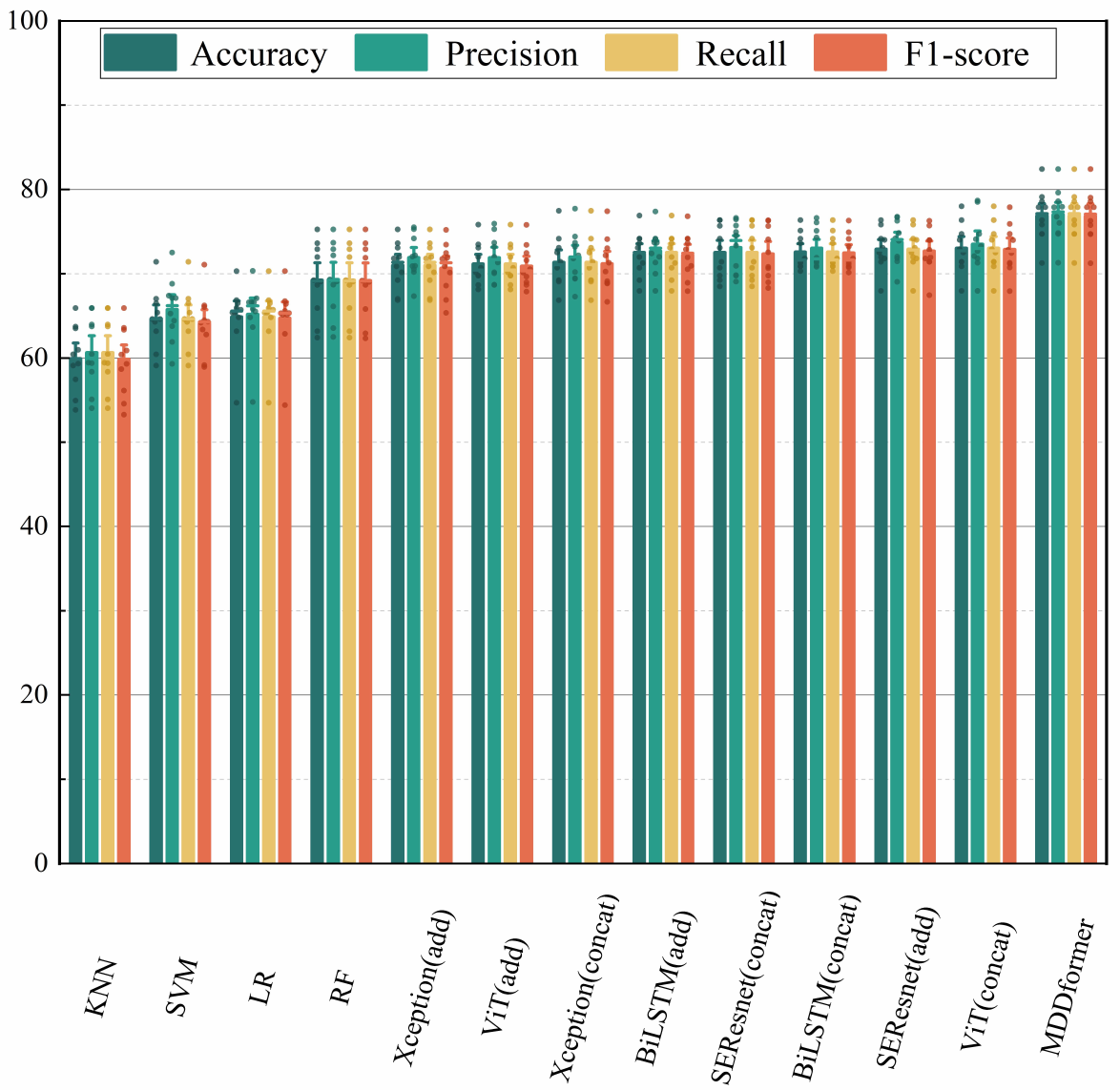}
	\caption{The grouped bar chart for the different baseline methods and MDDformer.} 
	\label{fig:group}
\end{figure}

Fig. \ref{fig:group} presents the classification performances using a bar chart. Each model is depicted by a group of four bars, each corresponding to one of the evaluation metrics. From left to right, the models are KNN, SVM, LR, RF, Xception (add), ViT (add), Xception (concat), BiLSTM (add), SEResnet (concat), BiLSTM (concat), SEResnet (add), ViT (concat), and the MDDformer. The MDDformer is represented by the last group of bars, displaying performance measures consistently around the 76-77\% mark for all four metrics. Notably, the performance of the MDDformer surpasses that of the other baseline models depicted in the chart.

\section{Conclusion}\label{sec:conclusion}

In this article, we collect a large-scale vlog dataset, i.e., the LMVD, for depression recognition among individuals navigating their daily lives based on audiovisual cues. The proposed dataset has 1823 samples (214 hours) from 1475 participants. To performance with the MDDformer, i.e., KNN, SVM, LR, RF, Xception, BiLSTM, SEResnet, and ViT. More importantly, our LMVD is the largest dataset for audiovisual depression recognition in individuals navigating their daily lives, which is a positive contribution to the affective computing field. In the future, we will augment the dataset and explore non-verbal behaviors for multimodal depression recognition.



\ifCLASSOPTIONcaptionsoff
  \newpage
\fi



%

\bibliographystyle{IEEEtran}
\bibliography{IEEEabrv,LMVD}
%








\begin{IEEEbiography}[{\includegraphics[width=1in,height=1.25in,clip,keepaspectratio]{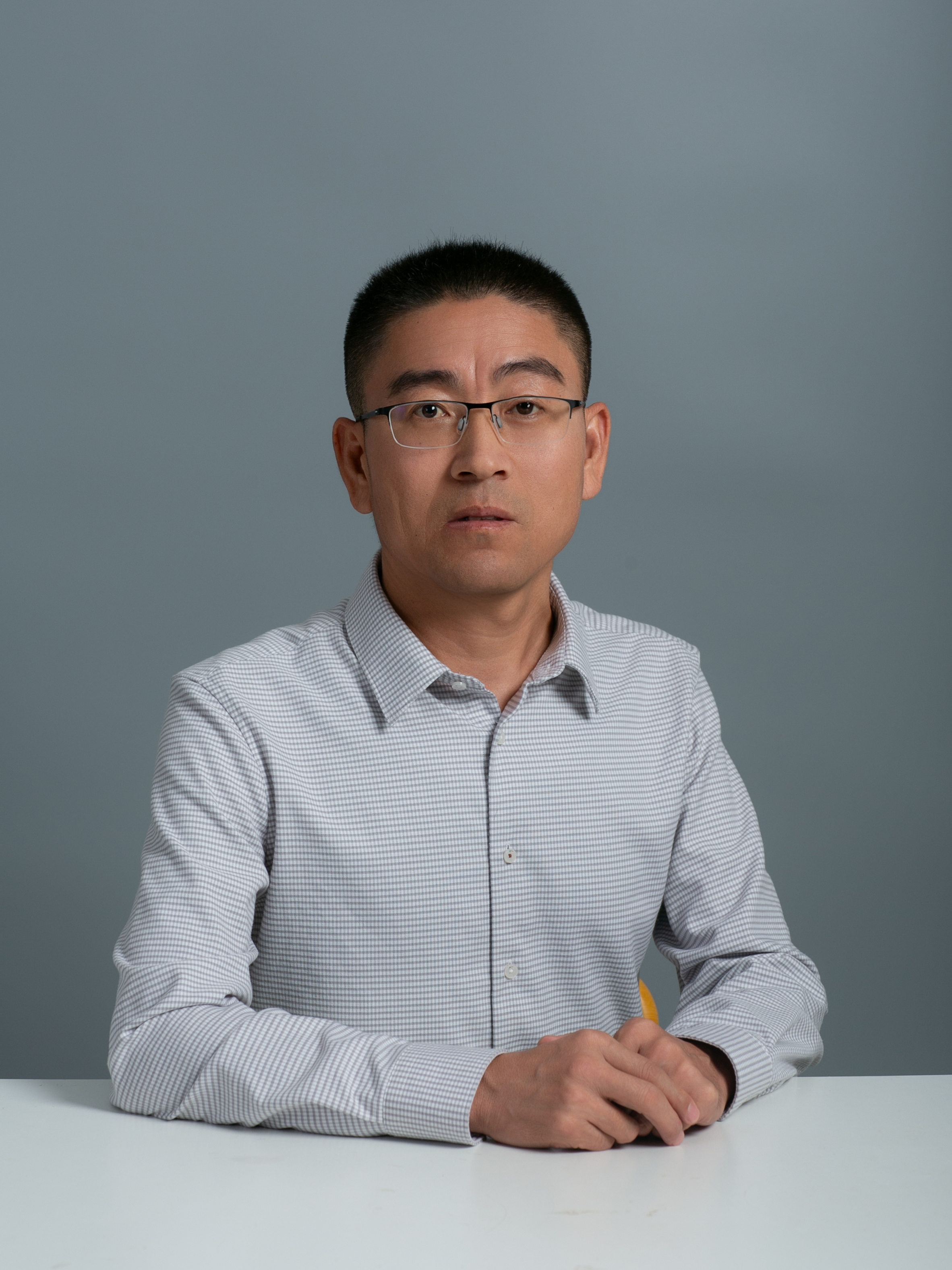}}]{Lang He}
received a Ph.D. degree in Computer Science and Technology from Northwest Polytechnical University (NPU), Xi’an, China, in 2019. From 2016 to 2017, he was a Visiting Scholar with Vrije Universiteit Brussel (VUB). He has been an associate professor with the School of Computer Science and Technology, Xi’an University of Posts and Telecommunications, Xi’an, China, since 2019. His current research interests include machine learning, multimodal depression recognition and analysis, emotion recognition, computer vision, etc. He is the first author and second author of the award-winning papers, for Audio Visual Emotion Challenge (AVEC2015) and AVEC2016.\end{IEEEbiography}

\begin{IEEEbiography}[{\includegraphics[width=1in,height=1.25in,clip,keepaspectratio]{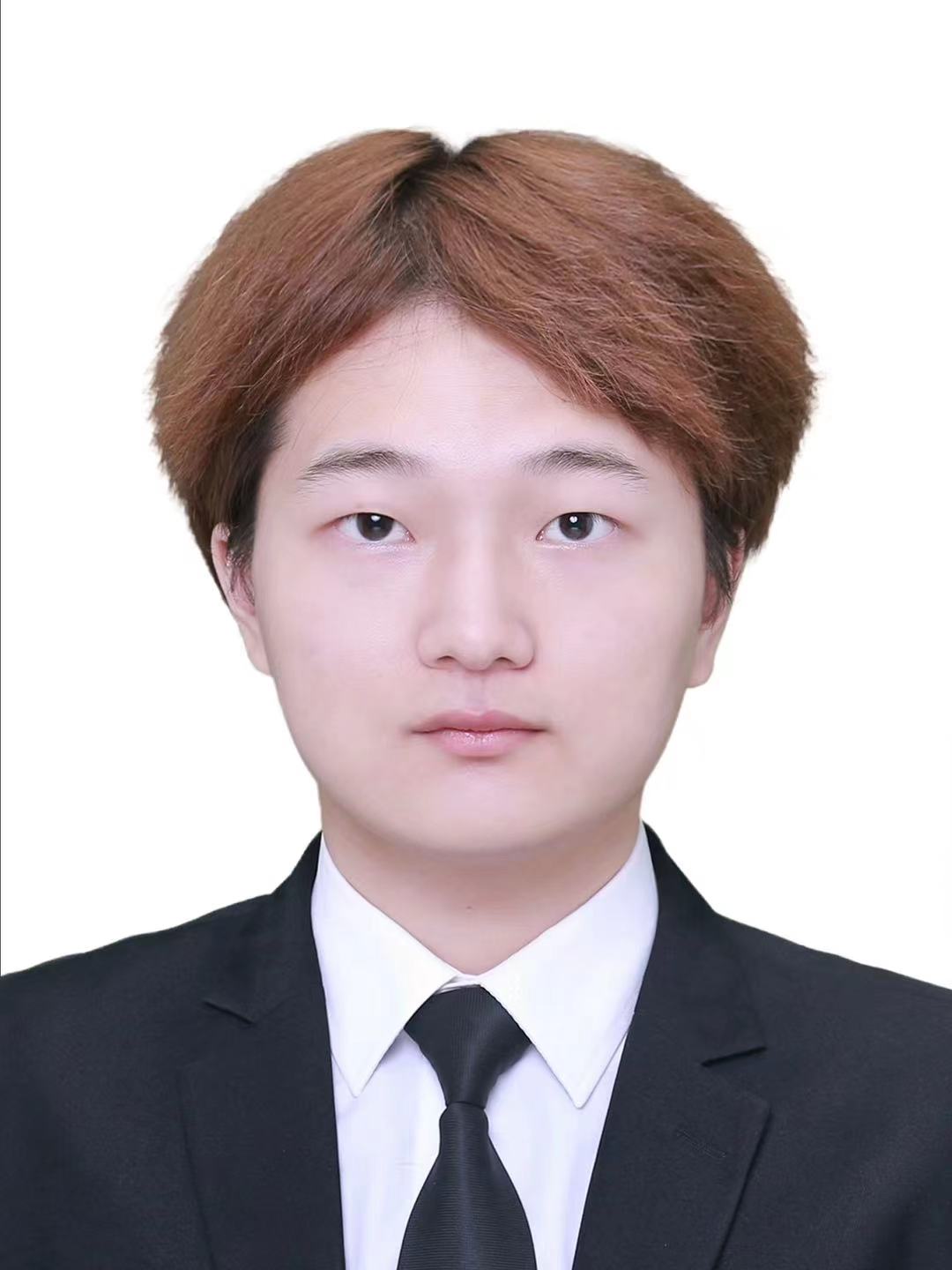}}]{Kai Chen}
	a master student at the School of Computer Science and Technology, Xi’an University of Posts and Telecommunications, Xi’an, China. His current research interest mainly focuses on multimodal depression recognition and analysis.
\end{IEEEbiography}

\begin{IEEEbiography}[{\includegraphics[width=1in,height=1.25in,clip,keepaspectratio]{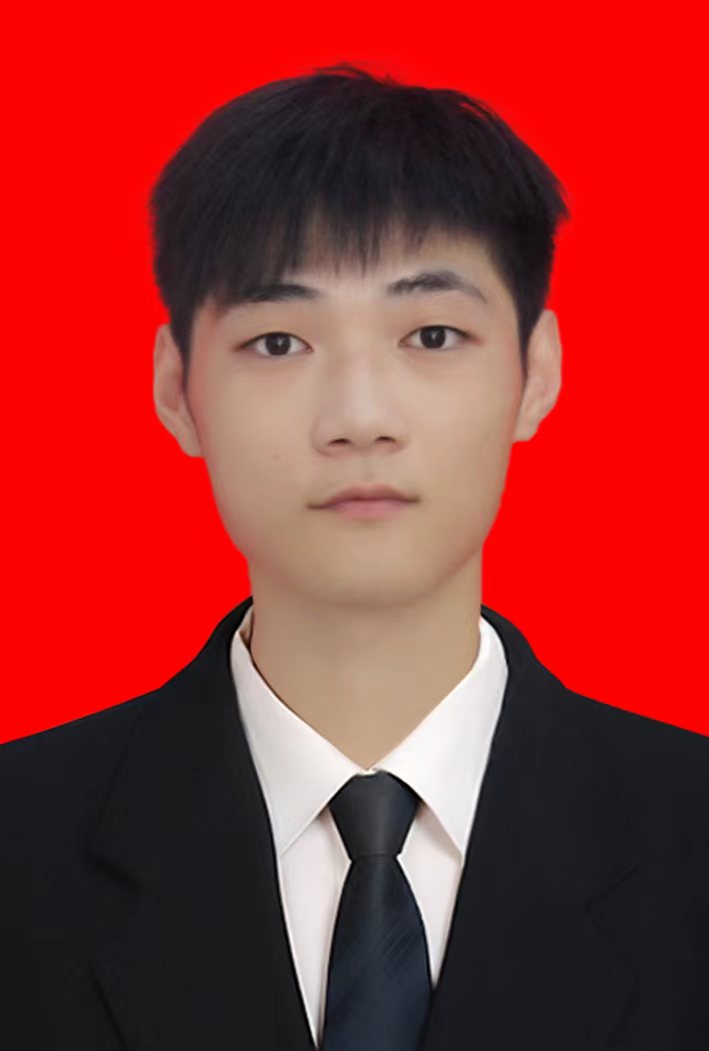}}]{Junnan Zhao}
	a master student at the School of Computer Science and Technology, Xi’an University of Posts and Telecommunications, Xi’an, China. His current research interest mainly focuses on multimodal depression recognition and analysis.
\end{IEEEbiography}
%
\begin{IEEEbiography}[{\includegraphics[width=1in,height=1.25in,clip,keepaspectratio]{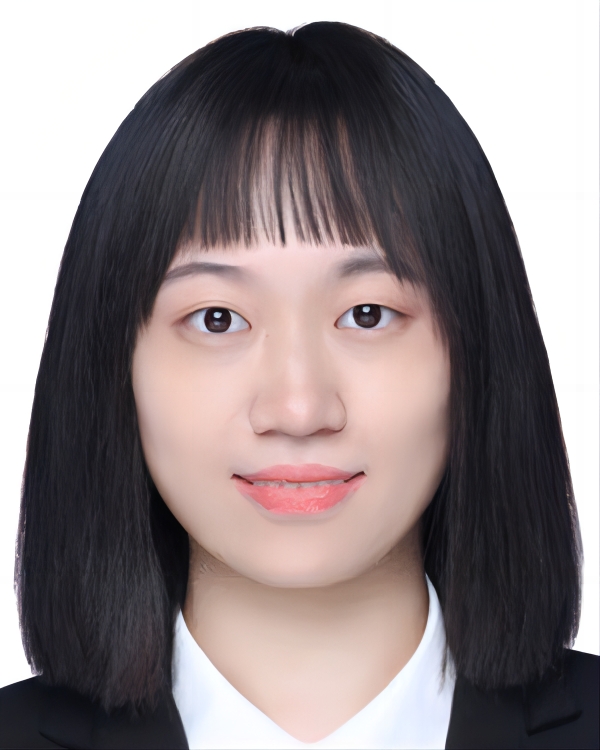}}]{Yimeng Wang}
	received the Ph.D. degree from the Department of Computer Science and Technology, Xi’an Jiaotong University, China, in 2022.  She is currently an Assistant Professor with the School of Computer Science and Technology, Xi’an University of Posts and Telecommunications, Xi’an. Her research interests include edge computing, industrial intelligence and system optimization.
\end{IEEEbiography}

\begin{IEEEbiography}[{\includegraphics[width=1in,height=1.25in,clip,keepaspectratio]{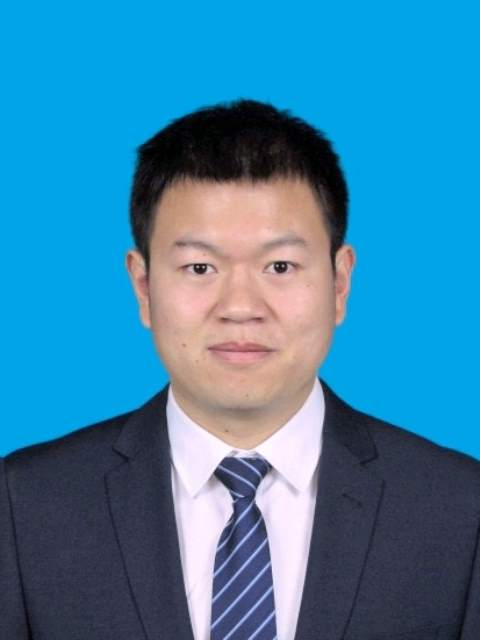}}]{Ercheng Pei}
	received his Master degree in Computer Application Technology and Ph.D degree in Computer Science and Technology from the Northwestern Polytechnical University (NPU), Xi'an, China in 2015 and 2020, respectively. Since 2021 he has been a lecturer at the School of Computer Science \& Technology, Xi'an University of Posts \& Telecommunications. His current research interest focuses on facial affective analysis, affective computing, computer vision and machine learning.
\end{IEEEbiography}

\begin{IEEEbiography}[{\includegraphics[width=1in,height=1.25in,clip,keepaspectratio]{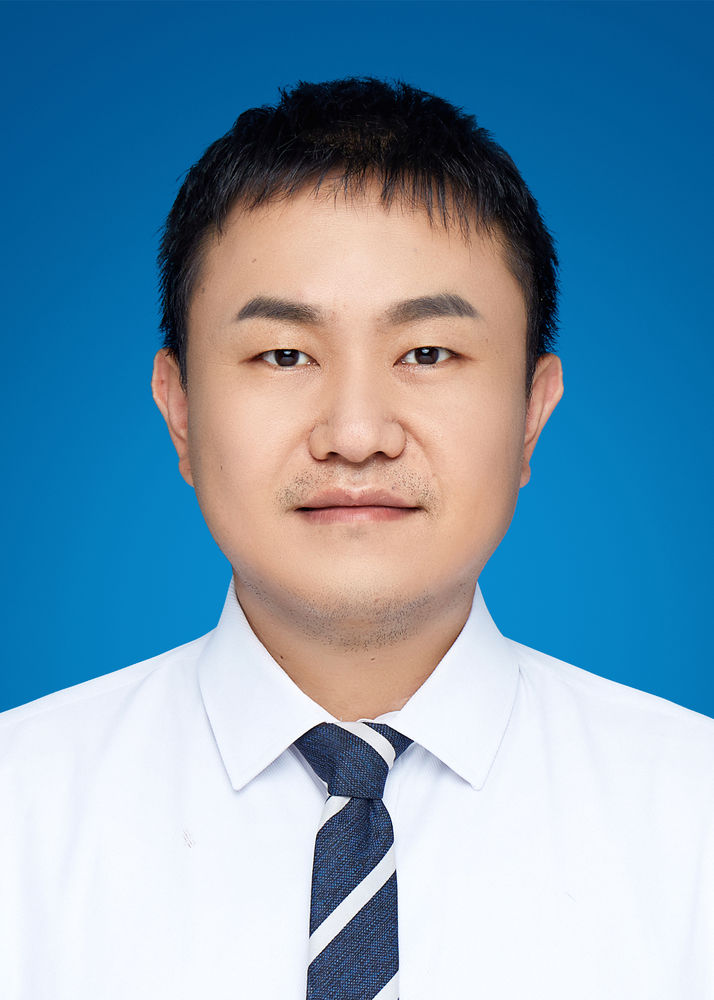}}]{Haifeng Chen}
	received the PhD degree in Computer Science and Technology from the Northwestern Polytechnical University (NPU), Xi'an, China in 2022, and he is currently a lecturer with the School of Electronic Information and Artificial Intelligence, Shaanxi University of Science and Technology since February. His research interests include affective computing, facial action unit detection, depression detection.
\end{IEEEbiography}

\begin{IEEEbiography}[{\includegraphics[width=1in,height=1.25in,clip,keepaspectratio]{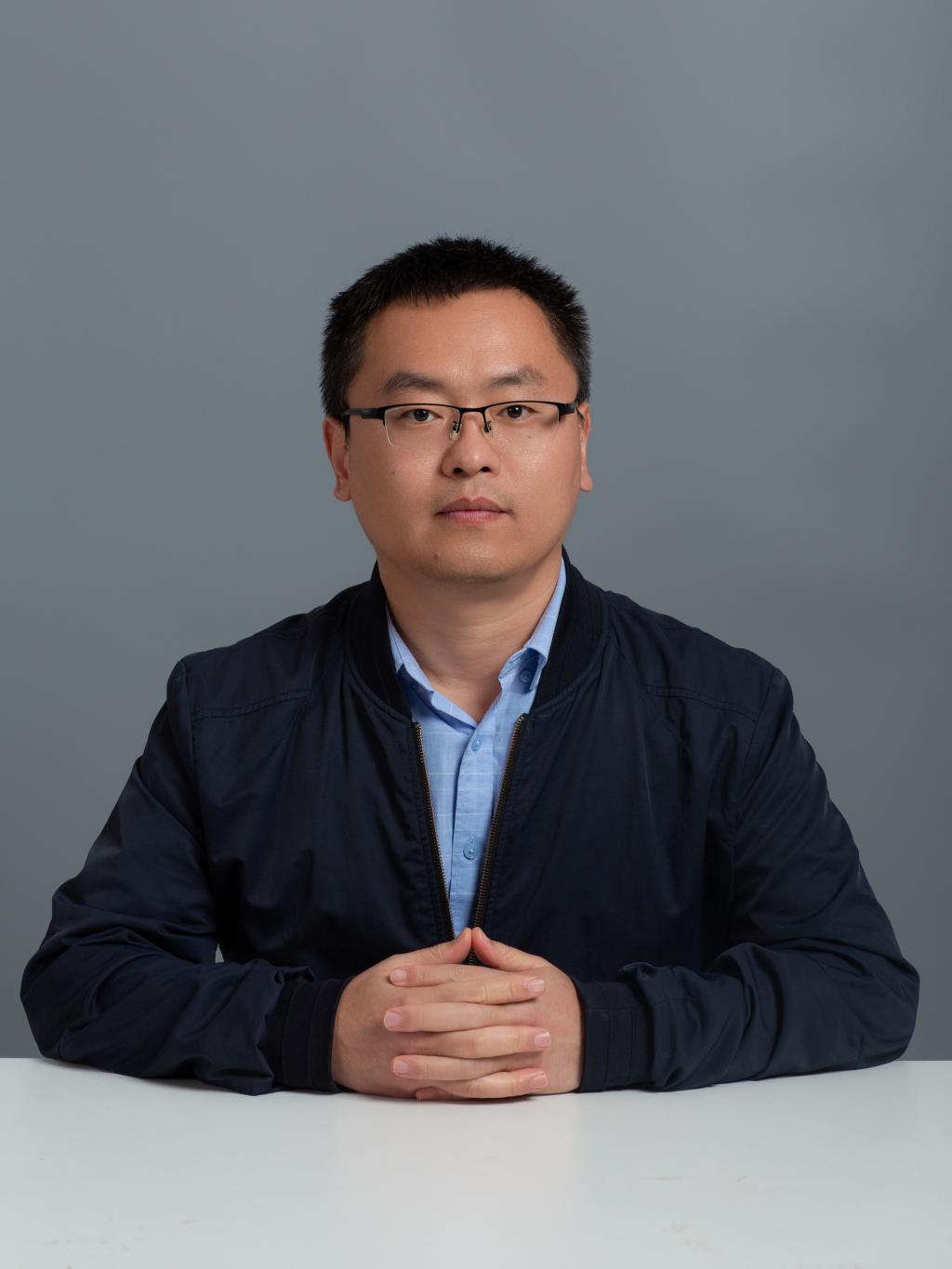}}]{Jiewei Jiang}
received the B.S. degree in communication engineering from the Henan University, Kaifeng, China, in 2006, the M.S. degree in communication and information system from the Xi'an University of Posts and Telecommunications, Xi’an, China, in 2009, and the Ph.D. degree in computer science and technology from the Xidian University, Xi'an, China, in 2018. He is currently an associate professor with the School of Electronic Engineering, Xi'an University of Posts and Telecommunications. His current research interests include deep learning, medical imaging processing, and artificial intelligence in medicine and industry.
\end{IEEEbiography}

\begin{IEEEbiography}[{\includegraphics[width=1in,height=1.25in,clip,keepaspectratio]{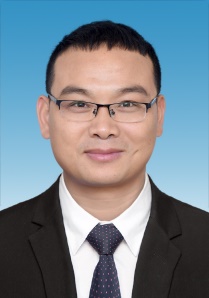}}]{Shiqing Zhang}
 (Member, IEEE) received the Ph.D. degree at school of Communication and Information Engineering, University of Electronic Science and Technology of China, in 2012. He was a postdoctor with the School of Electronic Engineering and Computer Science, Peking University, Beijing, China. Currently, he is a professor of the Institute of Intelligent Information Processing, Taizhou University, China. His research interests include affective computing and pattern recognition. He has published over 70 papers in journals and conferences such as IEEE Transactions on Affective Computing, IEEE Transactions on Multimedia, IEEE Transactions on Circuits and Systems for Video Technology, and ACM MM. He serves as an Associate Editor for IEEE Transactions on Affective Computing.

\end{IEEEbiography}

\begin{IEEEbiography}[{\includegraphics[width=1in,height=1.25in,clip,keepaspectratio]{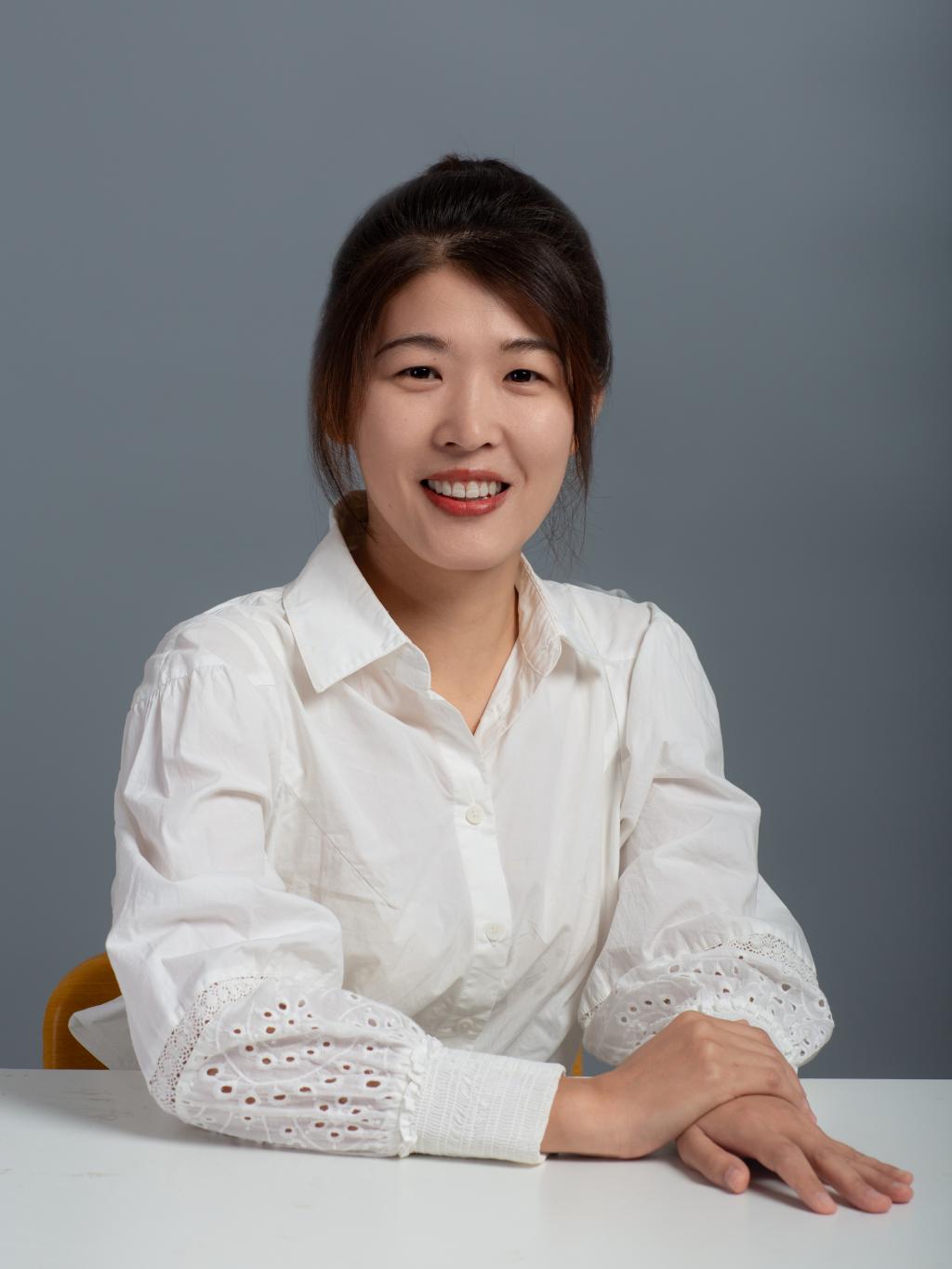}}]{Jie Zhang}
is currently an associate professor with the School of Computer Science and Technology, Xi’an University of Posts and Telecommunications. Her main research directions are intelligent wireless network, intelligent Internet of things, wireless intelligent perception, machine learning and mobile computing.
\end{IEEEbiography}

\begin{IEEEbiography}[{\includegraphics[width=1in,height=1.25in,clip,keepaspectratio]{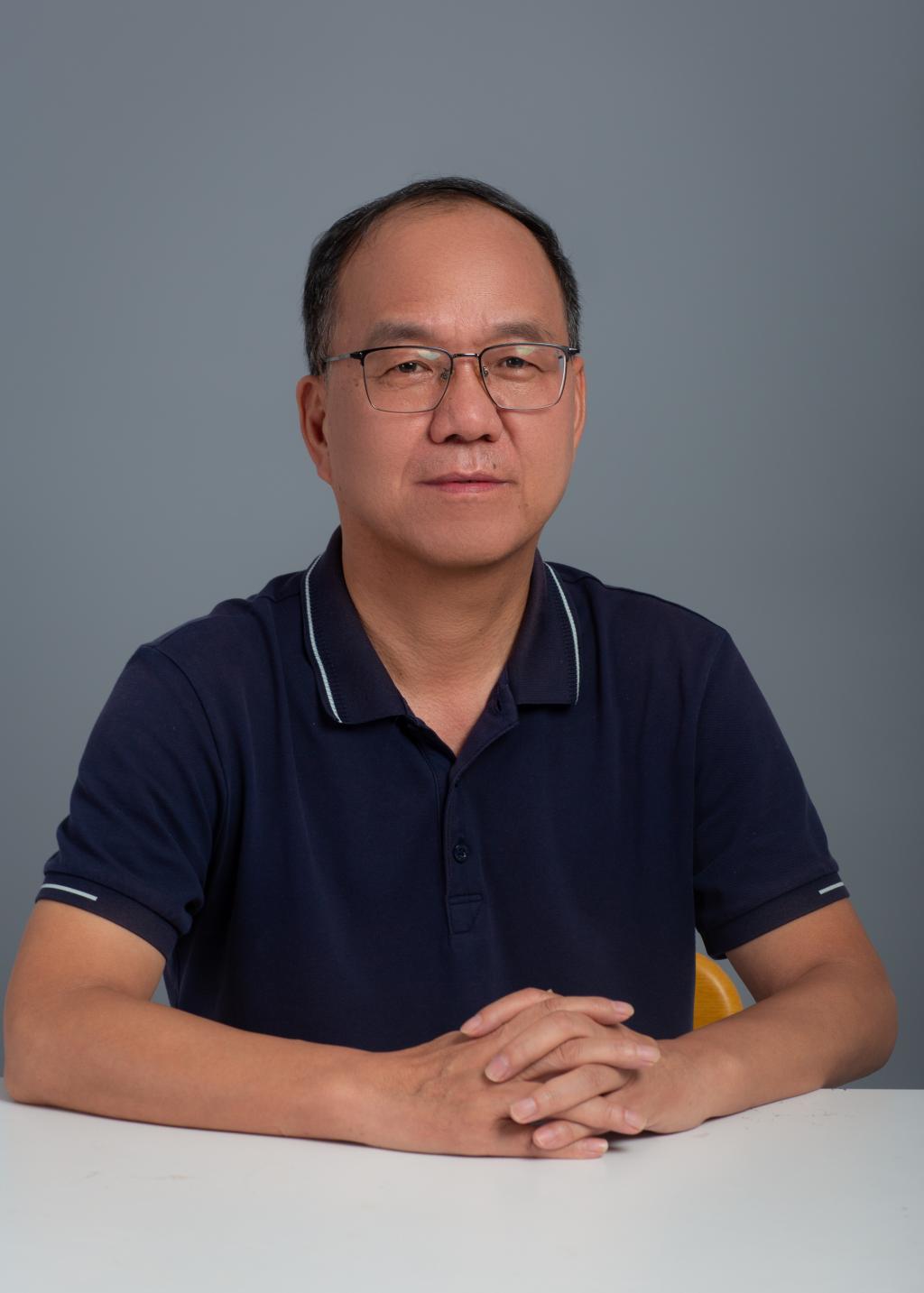}}]{Zhongmin Wang}
received his M.S. degree in mechatronic
engineering in 1993, and Ph.D. degree in
Mechanical Manufacture and Automation in 2000 from
Beijing Institute of Technology, Beijing, China. He was
with Xidian University from April 1993 to August 1997.
From February 2004 to February 2005, he was a visiting
scholarship at the robotics laboratory of the Australian
national university, Canberra, Australia. Since 2000, he
has been a full professor with the School of computer
science and technology at Xi’an University of Posts and
Telecommunications. His current research interests
include embedded intelligent perception, intelligent
information processing, machine learning, brain-computer interface and affective computing.

\end{IEEEbiography}

\begin{IEEEbiography}[{\includegraphics[width=1in,height=1.25in,clip,keepaspectratio]{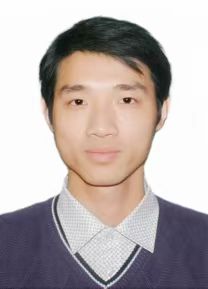}}]{Tao He}
is an associate professor in the College of Education at  Shenzhen University,his current research interests include learning analytics, cognitive computing, open education resources.

\end{IEEEbiography}

\begin{IEEEbiography}[{\includegraphics[width=1in,height=1.25in,clip,keepaspectratio]{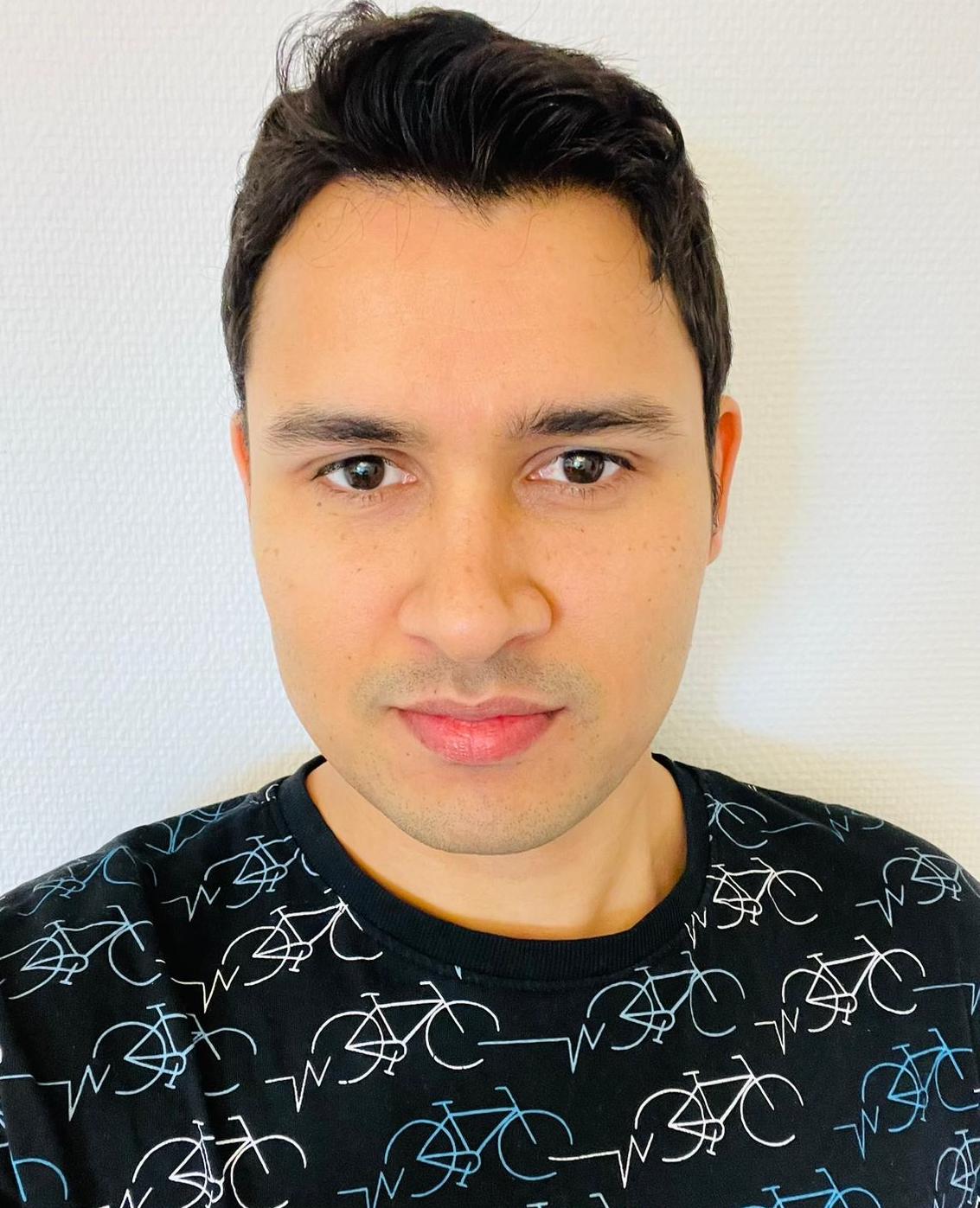}}]{Prayag Tiwari}
(Senior Member, IEEE) received the Ph.D. degree from the University of Padova, Padua, Italy. He is currently an Assistant Professor with Halmstad University, Halmstad, Sweden. He was a Postdoctoral Researcher with the Aalto University, Espoo, Finland and Marie Curie Researcher with the University of Padova. His name has appeared in the World's Top 2\% of Scientists released by Stanford University, Standard, CA, USA and Elsevier, Amsterdam, Netherlands. He was the recipient of the 2022 Best Paper Award for a paper published in Neural Networks, Elsevier. His papers were also the recipient of the Highly Cited Papers published in journals like Applied Sciences and Measurements. He has several publications in top journals and conferences, including Neural Networks, Journal of Physics A, Information Fusion, Knowledge-Based Systems, International Journal of Computer Vision, IEEE Transactions on Neural Networks and Learning Systems, IEEE Transactions on Fuzzy Systems Journal of Biomedical and Health Informatics, IEEE Transactions on Artificial Intelligence, IEEE Internet of Things Journal, IEEE BIBM, ACM Transactions On Internet Technology, ACM Computing Surveys, CIKM, SIGIR, AAAI, ECIR, ECML, and ACL. His research interests include artificial intelligence, quantum machine learning, deep learning, healthcare, bioinformatics, NLP, computer vision, intelligent System, and IoT.

\end{IEEEbiography}

\end{document}